\documentclass[10pt,twocolumn,letterpaper]{article}

\usepackage[pagenumbers]{cvpr} 

\usepackage{multicol, multirow}
\usepackage{pifont}
\usepackage[table]{xcolor}
\definecolor{lightblue}{RGB}{230, 240, 255}
\definecolor{blockA}{RGB}{230,240,255}  
\definecolor{blockB}{RGB}{240,255,240}  
\definecolor{blockC}{RGB}{255,245,230}  
\definecolor{cat1}{RGB}{235,245,255}  
\definecolor{cat2}{RGB}{240,255,240}  
\definecolor{cat3}{RGB}{255,245,220}  
\usepackage{makecell}
\usepackage[table]{xcolor}

\usepackage{xcolor}
\usepackage{soul}

\sethlcolor{lightblue}

\definecolor{cvprblue}{rgb}{0.21,0.49,0.74}
\usepackage[pagebackref,breaklinks,colorlinks,allcolors=cvprblue]{hyperref}

\def\paperID{662} 
\def\confName{CVPR}
\def\confYear{2026}

\title{MOS: Mitigating Optical-SAR Modality Gap for Cross-Modal Ship Re-Identification}

\author {
    Yujian Zhao\textsuperscript{1}, Hankun Liu\textsuperscript{2}, Guanglin Niu\textsuperscript{1\thanks{Corresponding author.}}
    \\
    \textsuperscript{1} School of Artificial Intelligence, Beihang University \\
    \textsuperscript{2} School of Computer Science and Engineering, Beihang University \\ 
    \{yjzhao1019, lhk2783497478, beihangngl\}@buaa.edu.cn 
}

\begin{document}
\maketitle
\begin{abstract}
Cross-modal ship re-identification (ReID) between optical and synthetic aperture radar (SAR) imagery has recently emerged as a critical yet underexplored task in maritime intelligence and surveillance. However, the substantial modality gap between optical and SAR images poses a major challenge for robust identification. To address this issue, we propose MOS, a novel framework designed to Mitigate the Optical–SAR modality gap and achieve modality-consistent feature learning for optical-SAR cross-modal ship ReID. MOS consists of two core components: (1) Modality-Consistent Representation Learning (MCRL) applies denoise SAR image procession and a class-wise modality alignment loss to align intra-identity feature distributions across modalities. (2) Cross-modal Data Generation and Feature fusion (CDGF) leverages a brownian bridge diffusion model to synthesize cross-modal samples, which are subsequently fused with original features during inference to enhance alignment and discriminability. Extensive experiments on the HOSS ReID dataset demonstrate that MOS significantly surpasses state-of-the-art methods across all evaluation protocols, achieving notable improvements of +3.0\%, +6.2\%, and +16.4\% in R1 accuracy under the \textit{ALL to ALL}, \textit{Optical to SAR}, and \textit{SAR to Optical} settings, respectively. The code and trained models will be released upon publication.
\end{abstract}    
\section{Introduction}
\label{sec:intro}
Ship re-identification (ReID) aims to identify the same ship across different images and is essential for maritime monitoring, tracking, and management. In maritime applications, ship imagery is typically acquired in both optical and synthetic aperture radar (SAR) modalities. SAR can capture high-resolution images under all-weather and day-night conditions, making it widely used for maritime surveillance. However, SAR images contain speckle noise and radar-specific textures, which challenge conventional feature extraction and matching. To address these issues, cross-modal ship ReID between optical and SAR imagery has recently emerged as a crucial task, aiming to match ships across heterogeneous sensors while bridging the large modality gap.  In recent years, although several works \cite{D2InterNet, MCL, CMShipReID, SMART-Ship, lu2025new} have provided valuable insights and achieved notable progress on this task, two major challenges still remain.
\begin{figure}[t]
  \centering
   \includegraphics[width=1.0\linewidth]{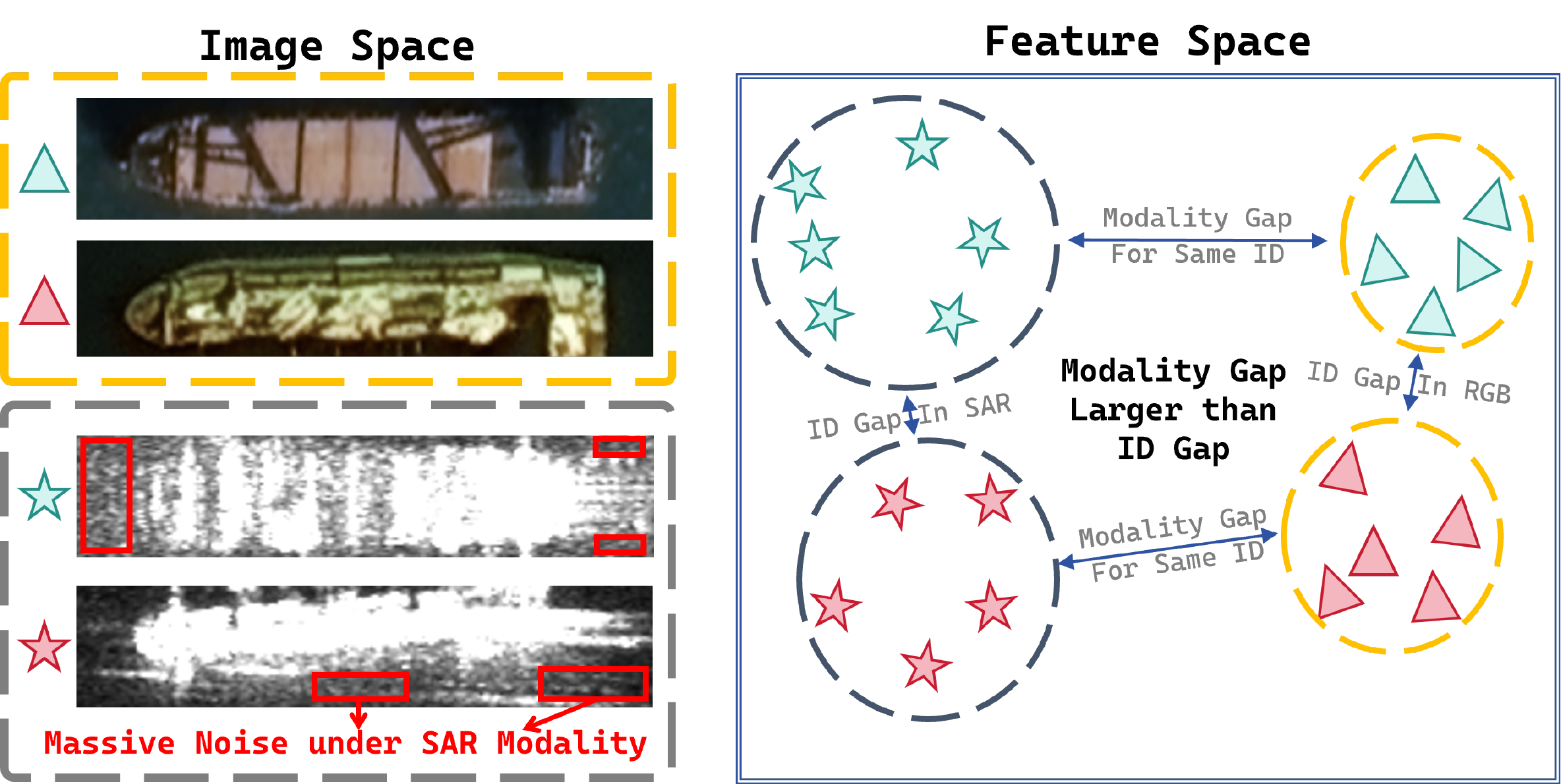}

   \caption{Noise of SAR image (left). Modality gap and ID gap in feature space (right). }
   \label{fig:introduction}
\end{figure}

\textbf{In cross-modal retrieval, the fundamental difference between optical and SAR imaging mechanisms leads to feature misalignment and a large modality gap.} In recent years, numerous efforts have been devoted to addressing this challenge. RQ-Net \cite{RQ-Net} constructs a robust quadratic network and a reconsider loss to match features from optical and SAR modalities. DEN \cite{den} designs a new fusion network structure called dual-encoder net to capture the detailed information and reduce the impact of modal differences. GaLD \cite{gald} introduces a cross-modal Gaussian localization distillation strategy to align the angle information between optical and SAR images. NSCF \cite{nscf} develops a content-filtering module to adaptively suppress regions that are difficult to align and enhance both coarse- and fine-grained features. RSENet \cite{RSENet} proposes a rotation and scale equivariant network to enforce model to accurately learn the information from two modalities. Despite these advances, most existing approaches rely on deep networks to implicitly learn high-dimensional feature representations, placing substantial demands on model capacity. To address this issue, we address this challenge by introducing Cross-modal Data Generation and Feature fusion (CDGF), a framework that synthesizes cross-modal samples and performs feature fusion across modalities. CDGF bridges the gap between optical and SAR imagery from both data and feature perspectives.

\textbf{In mixed-modality retrieval, modality differences frequently dominate identity differences, causing the model to favor same-modality samples, even if incorrect, while overlooking correct cross-modal matches.} SAPS \cite{SAPS} proposes a supervised alignment perception strategy to rectify spatial biases existing in SAR images and a modality difference selection module to compensate for the missing information in the optical images. Feng \etal~\cite{feng2023analytical} addresses the cognitive gap of CNNs between optical and SAR images by leveraging multi-order interactions to measure their representation capacity. Building upon prior research, we observe that the imaging mechanism of SAR inherently produces coherent speckle noise during the imaging process, which introduces significant interference in feature extraction. The left half of \cref{fig:introduction} illustrates the noise of SAR image. As the model ranks gallery samples by Euclidean or cosine distance during inference, it tends to focus on same-modality matches, as illustrated in the right part of \cref{fig:introduction}, limiting cross-modal generalization in real-world scenarios. To tackle this challenge, we propose a Modality-Consistent Representation Learning (MCRL) strategy, which denoises SAR images and incorporates a modality distance regularization term to further mitigate the modality gap, ultimately enhancing cross-modal recognition performance.

In summary, our main contributions are as follows:
\begin{itemize}
    \item We propose a novel framework MOS for cross-modal ship re-identification, which can accurately denoise SAR images and significantly mitigate optical-SAR modality gap throughout both the training and inference phases.
\end{itemize}
\begin{itemize}
    \item We design a Modality-Consistent Representation Learning (MCRL) strategy that can remove the inherent noise in SAR images and reduce the modal distance through our proposed class-wise modality alignment loss.
\end{itemize}
\begin{itemize}
    \item We construct a Cross-modal Data Generation and Feature Fusion (CDGF) method to synthesize cross-modal samples and fuse their features with the original features during inference.
\end{itemize}
\begin{itemize}
    \item Extensive experiments on the HOSS ReID dataset show our MOS achieves superior results, significantly surpassing SOTA methods. Detailed ablation studies further validate the exceptional benefits of our modules.
\end{itemize}

\section{Related work}
\subsection{Ship Re-identification}
Ship re-identification (ReID) \cite{ghahremani2019towards, qiao2020marine, D2InterNet, zwemer2021multi, MCL, CMShipReID} aims to match the same ship across non-overlapping cameras or heterogeneous sensors, which is essential for maritime surveillance and vessel tracking. While person and vehicle ReID \cite{pose2id,xu2025identity, sheng2023discriminative, he2024instruct, wang2024exploring, gao2023identity, warship,liu2025automatic, hao2025masked, liu2025try, 11209188, liu2025looking} have achieved remarkable progress through large-scale datasets and advanced deep feature learning, ship ReID remains far less explored due to the unique challenges of maritime environments. Liu \etal~\cite{D2InterNet} introduces a ship ReID dataset called ShipReID-2400 from a real-world intelligent waterway traffic monitorin system and a simple but strong baseline called D2InterNet to extract discriminative local features. Qiao \etal~\cite{qiao2020marine} creates an large-scale vessel retrieval dataset named VesselID-539 and describs a fusion-based multi-view architecture to combine global and fine-grained local feature. Zwemer \etal~\cite{zwemer2021multi} introduces a new Vessel-reID dataset and a multi-level contrastive learning method to deal with the challenges of positive sample selection in unsupervised training. Qian \etal~\cite{MCL} developes a multi-level contrastive learning framework for ship ReID and incorporate a novel contrastive loss for training. 
Xu \etal~\cite{CMShipReID} construct the new dataset called CMShipReID, which contains visible light, near-infrared, and thermal infrared modalities collected by autonomous aerial vehicle.

For ship Re-identification under the optical–SAR modality, related work remains scarce. To the best of our knowledge, only two works \cite{hoss, SMART-Ship} have specifically addressed this task to date. Wang \etal~\cite{hoss} pioneers this task by presenting the first dataset called HOSS ReID and proposes a baseline method, TransOSS, which is built on the Vision Transformer architecture. Fan \etal~\cite{SMART-Ship} constructs a comprehensive ship ReID dataset, SMART-Ship, covering five modalities: visible light, SAR, panchromatic, multi-spectral, and near-infrared. Since the latter has not been publicly released, we adopt TransOSS \cite{hoss} as the baseline and enhance it with our proposed MOS, achieving improved retrieval performance.

\subsection{Data generation and feature fusion}

With the rapid development of generative models \cite{goodfellow2020generative, croitoru2023diffusion, rombach2022high, ho2020denoising, li2025dual, gushchin2024adversarial}, image generation and translation have become increasingly important tasks in computer vision. These techniques enable the synthesis of new images, cross-domain translation, and augmentation of underrepresented modalities, providing effective solutions for bridging modality gaps. Yuan \etal~\cite{pose2id} proposes a feature centralization framework for person ReID, aggregating same-identity features generated by a stable diffusion model to enhance identity representation stability. Siddiqui \etal~\cite{siddiqui2025dlcr} constructs a data expansion framework called DLCR to synthesize cloth-changing person ReID data via diffusion and large language models, and introduces strategies to progressive learning these generated data to obtain more robust feature representation. Wu \etal~\cite{wu2024optical} introduces a novel cross-domain attention GAN network for optical-to-SAR image translation and utilize the generated samples can obtain higher performance. Ma \etal~\cite{ma2025multimodal} introduce a method for multimodal cluster analysis by fusing embeddings obtained from a variational autoencoder \cite{kingma2013auto}. The aforementioned works provide valuable insights into feature fusion from various domains and perspectives. At the same time, a lot of works\cite{bai2023conditional, guo2024scene, he2025dogan, zhao2024hvt} have been dedicated to tackling the problem of optical-to-SAR image translation. He \etal~\cite{he2025dogan} proposes DOGAN, a distillation with no labels-based optical-prior
driven generative adversarial network, to transfer SAR images to optical modality.  Wang \etal~\cite{wang2023optical} utilize denoising diffusion implicit model \cite{song2020denoising} to generate inverse SAR images from the synthesized optical counterparts. Inspired by the aforementioned works, we employ a Brownian Bridge diffusion model \cite{li2023bbdm} during inference to generate SAR samples from optical inputs. Both the original and generated samples are fed into the model to extract high-dimensional features, which are then fused to produce modality-complementary representations, thereby improving cross-modal retrieval accuracy.
\section{Methodology}
\begin{figure*}[t]
  \centering
   \includegraphics[width=1.0\linewidth]{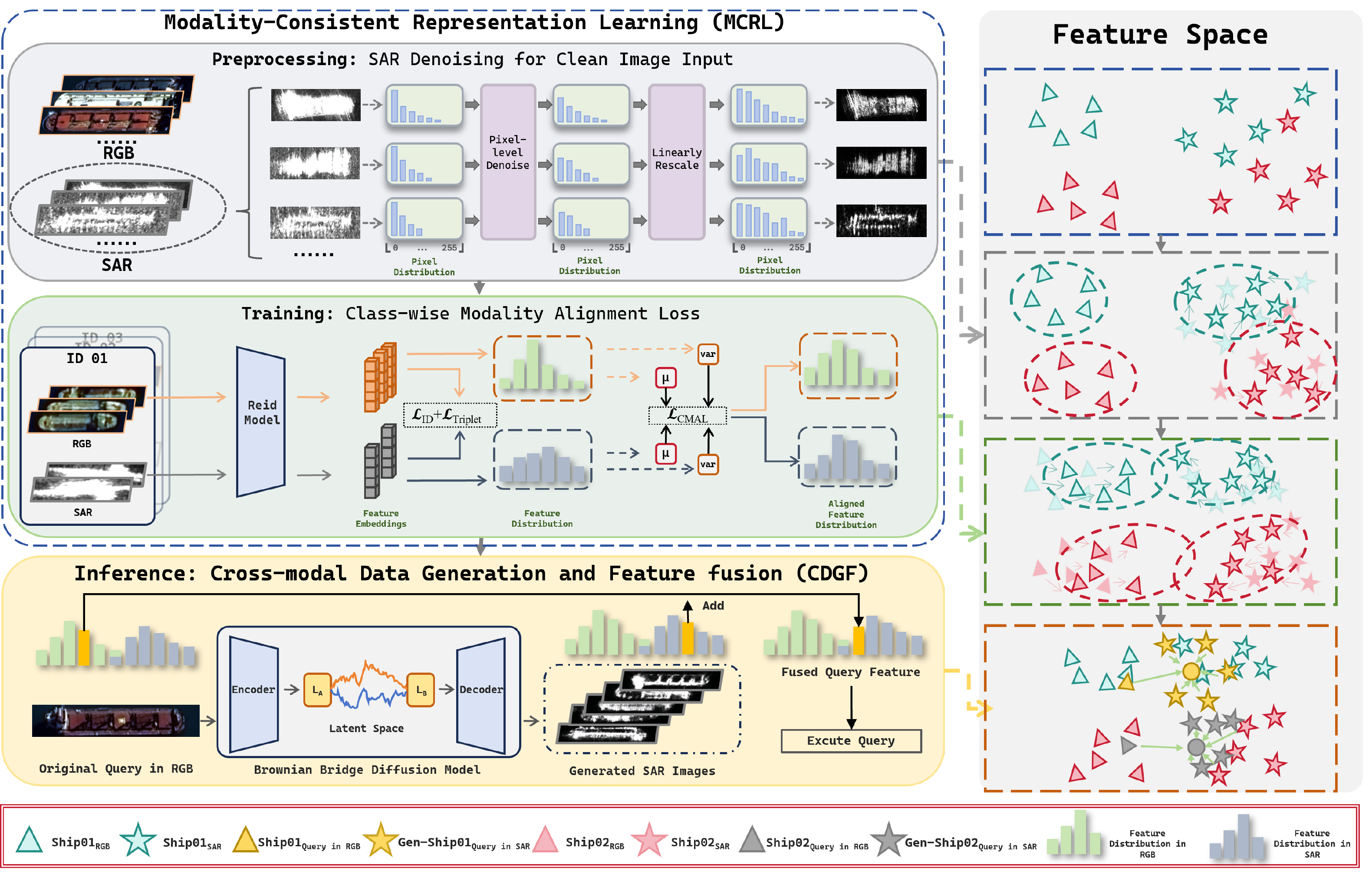}

   \caption{Overview of our proposed MOS. MOS consists of two parts: Modality-Consistent Representation Learning (MCRL) and Cross-modal Data Generation and Feature fusion (CDGF). MCRL performs effective noise suppression on SAR images and introduces a class-wise modality alignment loss to align the distributions of multi-modal samples sharing the same identity during training. CDGF, on the other hand, generates corresponding SAR samples from optical inputs during inference and fuses them with original features to further enhance retrieval performance.}
   \label{fig:overview}
\end{figure*}
\subsection{Overview}

The optical–SAR ship ReID dataset is denoted as $\mathcal{D} = \{(I_i, y_i, m_i)\}_{i=1}^{N}$, where $I_i$ represents the $i$-th ship image, $y_i \in \{1, \dots, C\}$ denotes the ship identity label, and $m_i \in \{\mathrm{opt}, \mathrm{sar}\}$ indicates the modality, i.e., optical or SAR. The dataset is split into disjoint training and testing subsets, $\mathcal{D}_{\mathrm{train}}$ and $\mathcal{D}_{\mathrm{test}}$, with non-overlapping identities. Each image $I$ is transformed into a $d$-dimensional embedding vector through a feature extractor $f_\theta(\cdot)$ parameterized by $\theta$:
\begin{equation}
\mathbf{z} = f_\theta(I) \in \mathbb{R}^d,
\end{equation}
The goal of cross-modal ship ReID is to learn a modality-invariant embedding function $f_\theta(\cdot)$ that enables accurate retrieval on the test set $\mathcal{D}_{test}$. 
Given a query image $I_q$ and a gallery $\mathcal{G} = \{I_g\}$, the retrieval target is defined as
\begin{equation}
I_g^* = \arg\min_{I_g \in \mathcal{G}} \mathrm{dist}(f_\theta(I_q), f_\theta(I_g)),
\end{equation}
where $\mathrm{dist}(\cdot,\cdot)$ denotes a distance metric such as Euclidean or cosine distance. 
Ideally, the selected $I_g^*$ should share the same identity with the query $I_q$ regardless of modality differences.

To achieve this, the objective of the optical–SAR ship ReID task is to learn the optimal parameters $\theta^\ast$ of the feature extractor $f_\theta(\cdot)$ that yield the highest recognition accuracy on the test set. Formally,
\begin{equation}
\theta^\ast = \arg\max_{\theta} \; \mathrm{Acc}\big(f_\theta; \mathcal{D}_{\mathrm{test}}\big),
\end{equation}
where $\mathrm{Acc}(\cdot)$ denotes the overall identification accuracy across modalities. 

To better optimize the feature extractor and obtain a more discriminative parameter set $\theta$, we propose a novel framework named MOS to Mitigate Optical-SAR modality gap. As illustrated in \cref{fig:overview}, MOS consists of two complementary components: the Modality-Consistent Representation Learning (MCRL) module and the Cross-modal Dats Generation and Feature fusion (CDGF) module. 
Specifically, MCRL focuses on aligning the feature distributions of different modalities through a shared embedding space, while CDGF leverages a brownian bridge diffusion model to generate cross-modal samples and fuse features of original and generated samples. Together, these two modules collaboratively enhance the model’s capability to learn modality-invariant yet identity-discriminative representations, thus improving optical–SAR ship ReID performance.

\subsection{Modality-consistent representation learning}
\label{sec3.2}

Optical and SAR images exhibit significant appearance discrepancies due to their inherently different imaging mechanisms, leading to a large modality gap in feature space. Moreover, SAR images are often corrupted by coherent speckle noise, which further deteriorates the robustness of feature extraction. 

\textbf{Denoise SAR image.} Given a SAR sample $(I_i, y_i, \mathrm{sar})$, where the image $I_i$ has spatial resolution $H \times W$. We observe that noise frequently occurs in areas with low pixel values, so we first arrange all pixels in ascending order:
\begin{equation}
\begin{split}
\{p_1, p_2, \dots, p_{HW}\}_{\mathrm{sorted}} & = 
\mathrm{sort}\big(\{I_i(m,n) \mid  \\ 1 \le m \le H, 
&\quad 1 \le n \le W\}\big),
\end{split}
\end{equation}
where $I_i(m,n)$ denotes the pixel value at spatial position $(m,n)$.
To mitigate speckle noise in SAR images, the lowest $\alpha\%$ of pixel values, which predominantly correspond to noise, are discarded. 
The remaining top $(1-\alpha)\%$ pixel values are subsequently linearly rescaled to the standard 8-bit range $[0,255]$:

\begin{equation}
\hat{p}_k = \frac{255 (p_k - p_{\min})}{p_{\max} - p_{\min} + \epsilon}, 
\quad k \in \mathcal{K}_\alpha
\end{equation}
where $p_{\min}$ and $p_{\max}$ denote the minimum and maximum pixel values among the retained subset, and $\mathcal{K}_\alpha = \mathrm{Top}_{1-\alpha\%}(\{p_i\}_{\mathrm{sorted}})$ denotes the indices of the top $(1-\alpha\%)$ percentile values in the  $\{p_i\}_{sorted}$. 
$\epsilon$ is a small constant introduced for numerical stability, 
and $\hat{p}_k$ represents the denoised and rescaled pixel intensity.

\textbf{Class-wise Modality Alignment Loss.} Although denoising contributes to the acquisition of more precise features, a substantial modality gap still exists in the feature space due to the inherent differences between optical and SAR imaging mechanisms. To further mitigate modality gap, we introduce a class-wise modality alignment loss that constrains the class-wise feature alignment between modalities.


For a batch of $B$ samples $\{(I_i, y_i, m_i),i=1, \dots, B\}$ with feature embeddings $F = \{f_1, f_2, \dots, f_B\}$, our goal is to aggregate samples with the same ID and separate samples with different IDs. Consider the class-conditional feature distributions for a given identity $c$ under the optical and SAR modalities:
$p_{\mathrm{opt}}^c,\quad p_{\mathrm{sar}}^c$. Under the Gaussian approximation \(p_{\mathrm{opt}}^c=\mathcal{N}(\mu_{\mathrm{opt}}^c,\Sigma_{\mathrm{opt}}^c)\) and \(p_{\mathrm{sar}}^c=\mathcal{N}(\mu_{\mathrm{sar}}^c,\Sigma_{\mathrm{sar}}^c)\), the squared Wasserstein-2 distance admits a closed form:
\begin{equation}
\begin{split}
&W_2^2\big(p_{\mathrm{opt}}^c,p_{\mathrm{sar}}^c\big) = \big\lVert \mu_{\mathrm{opt}}^c - \mu_{\mathrm{sar}}^c \big\rVert_2^2 \\
&\quad + \mathrm{Tr}\!\Big(\Sigma_{\mathrm{opt}}^c + \Sigma_{\mathrm{sar}}^c - 2\big(\Sigma_{\mathrm{opt}}^{c\,1/2}\Sigma_{\mathrm{sar}}^c\Sigma_{\mathrm{opt}}^{c\,1/2}\big)^{1/2}\Big).
\end{split}
\label{eq:w2_exact}
\end{equation}
The first term quantifies the disparity between the means of the two modalities, while the second term measures the divergence of their covariances. Directly minimizing the covariance term in Eq.~\eqref{eq:w2_exact} requires matrix square roots and is computationally expensive. At the same time, it is noted that in high-dimensional feature spaces, off-diagonal correlations between different feature dimensions are often weak or noisy. Under the diagonal-covariance approximation $\Sigma_{\cdot}^c \approx \mathrm{diag}((\sigma_{\cdot}^c)^2)$, the covariance contribution in Eq.~\eqref{eq:w2_exact} simplifies to the squared Euclidean distance between per-dimension standard deviations:
\begin{equation}
\begin{split}
&\mathrm{Tr}\Big(\Sigma_{\mathrm{opt}}^c + \Sigma_{\mathrm{sar}}^c 
- 2(\Sigma_{\mathrm{opt}}^{c\,1/2}\Sigma_{\mathrm{sar}}^c\Sigma_{\mathrm{opt}}^{c\,1/2})^{1/2}\Big) \\
& \approx \|\sigma_{\mathrm{opt}}^c - \sigma_{\mathrm{sar}}^c\|_2^2,
\end{split}
\end{equation}
where $\sigma_{\cdot}^c=\sqrt{\mathrm{diag}(\Sigma_{\cdot}^c)}$ denotes the per-dimension standard deviation vector. Empirically, such diagonal approximations have been shown to retain most of the alignment effect of the full Wasserstein-2 distance in high-dimensional embeddings while providing a robust and tractable regularization term.

For each identity $c$, we define the subsets of samples from the optical and SAR modalities as
\begin{equation}
\begin{split}
\mathcal{O}_c = \{ (I_i, y_i, m_i) \mid y_i = c,\, m_i = \mathrm{opt} \}, \\
\mathcal{S}_c = \{ (I_i, y_i, m_i) \mid y_i = c,\, m_i = \mathrm{sar} \}.
\end{split}
\end{equation}
We only consider the identities that appear in both modalities, forming the valid identity set
\begin{equation}
\mathcal{C} = \{ c \mid |\mathcal{O}_c| > 0 \ \text{and}\  |\mathcal{S}_c| > 0 \}.
\end{equation}

For each identity $c \in \mathcal{C}$, the class-wise mean vectors and element-wise variances are computed as
\begin{equation}
\begin{split}
\mu_{\mathrm{opt}}^c = \frac{1}{|\mathcal{O}_c|} \sum_{(I_i,y_i,m_i)\in \mathcal{O}_c} f_i, \\
\mu_{\mathrm{sar}}^c = \frac{1}{|\mathcal{S}_c|} \sum_{(I_i,y_i,m_i)\in \mathcal{S}_c} f_i, 
\end{split}
\end{equation}
\begin{equation}
\begin{split}
\mathrm{var}_{\mathrm{opt}}^c = \frac{1}{|\mathcal{O}_c|} \sum_{(I_i,y_i,m_i)\in \mathcal{O}_c} (f_i - \mu_{\mathrm{opt}}^c) \odot (f_i - \mu_{\mathrm{opt}}^c), \\
\mathrm{var}_{\mathrm{sar}}^c = \frac{1}{|\mathcal{S}_c|} \sum_{(I_i,y_i,m_i)\in \mathcal{S}_c} (f_i - \mu_{\mathrm{sar}}^c) \odot (f_i - \mu_{\mathrm{sar}}^c),
\end{split}
\end{equation}
where $\odot$ denotes element-wise multiplication.

Based on these statistics, the class-wise modality alignment loss is formulated as
\begin{equation}
\begin{split}
\mathcal{L}_{\mathrm{CMAL}} = \frac{1}{|\mathcal{C}|} \sum_{c \in \mathcal{C}} \bigg(
& \|\mu_{\mathrm{opt}}^c - \mu_{\mathrm{sar}}^c\|_2^2 \\
& + \|\mathrm{var}_{\mathrm{opt}}^c - \mathrm{var}_{\mathrm{sar}}^c\|_2^2 \bigg),
\end{split}
\label{cmal}
\end{equation}
The first term in Eq.~\eqref{cmal} minimizes the distance between the optical and SAR class centroids and reducs the modality gap at the class level. The second term encourages consistency in the intra-class feature dispersions between modalities. Together, these two components provide a tractable approximation of the Wasserstein-2 distance, enforcing both mean and variance consistency in the embedding space.

\textbf{Overall loss function.} The proposed CMAL can be jointly optimized with standard identity classification and triplet losses:
\begin{equation}
\mathcal{L} = \lambda_\mathrm{id}\mathcal{L}_{\mathrm{ID}} + \lambda_{\mathrm{tri}} \mathcal{L}_{\mathrm{Triplet}} + \lambda_{\mathrm{cmal}} \mathcal{L}_{\mathrm{CMAL}},
\end{equation}
where $\mathcal{L}_{\mathrm{id}}$ denotes the standard identity cross-entropy loss, $\mathcal{L}_{\mathrm{tri}}$ represents the triplet loss \cite{hermans2017defense} that enforces relative feature distances, and $\mathcal{L}_{\mathrm{CMAL}}$ is the proposed CMAL that minimizes the feature discrepancy between optical and SAR modalities. Following the baseline TransOSS \cite{hoss} configuration, we set $\lambda_{id} = \lambda_{\mathrm{tri}} = 1$.

\begin{table*}
  \caption{Comparisons between the proposed MOS and some state-of-the-art methods on the HOSS ReID datasets. \textbf{Bold} values indicate the best performance, and the second-best results are \underline{underlined}.}
  \label{results}
  \centering
  \setlength{\tabcolsep}{3pt}
  \begin{tabular}{ccc|cccc|cccc|cccc}
    \toprule
    \multirow{2}{*}{Task type} & \multirow{2}{*}{Method} & \multirow{2}{*}{Venue}  & \multicolumn{4}{c|}{\textit{ALL to ALL}}  & \multicolumn{4}{c|}{\textit{Optical to SAR}} & \multicolumn{4}{c}{\textit{SAR to Optical}} \\
    & & & mAP & R1 & R5 & R10 & mAP & R1 & R5 & R10 & mAP & R1 & R5 & R10 \\ 
    \hline
    \rowcolor{cat1}\multicolumn{15}{l}{General vision methods } \\
    \hline
    
    \rowcolor{cat1} & ViT-base \cite{vit} & Arxiv 2020 & 43.0 & 56.2 & 64.8 & 69.9 & 21.5 & 12.3 & 33.8 & 55.4 & 17.9 & 10.4 & 25.4 & 32.8\\
    \rowcolor{cat1} \multirow{-2}{*}{-} & DeiT-base \cite{Deit} & ICML 2021& 47.2 & 58.1 & 69.6 & 74.1 & 25.9 & 16.1 & 36.7 & 59.1 & 26.1 & 10.3 & 35.7 & 52.0\\
     \hline
    \rowcolor{cat2}\multicolumn{15}{l}{Single modality ReID methods } \\
    \hline
    \rowcolor{cat2} & AGW \cite{AGW} & TPAMI 2021 & 43.6 & 57.4 & 64.2 & 68.8 & 17.2 & 7.7 & 29.2 & 38.5 & 21.1 & 14.9 & 34.3 & 46.3\\
    \rowcolor{cat2} & TransReID \cite{transreid} & ICCV 2021 & 48.1 & 60.8 & 69.3 & 73.9 & 27.3 & 18.5 & 40.0 & 58.5 & 20.9 & 11.9 & 34.3 & 43.3 \\
    \rowcolor{cat2} \multirow{-3}{*}{\cellcolor{cat2}Person ReID}  & SOLIDER \cite{SOLIDER} & CVPR 2023 & 38.2 & 50.6 & 63.1 & 69.9 & 23.1 & 12.3 & 38.5 & 52.3 & 14.6 & 10.4 & 16.4 & 31.3 \\
    \hline
    \rowcolor{cat2} Ship ReID & D2InterNet \cite{D2InterNet} & SIGIR 2025 & 50.2 & 59.1 & 71.6 & 79.0 & 33.0 & 21.5 & 41.5 & 69.8 & 28.8 & 25.4 & 38.8 & 50.7 \\
    \hline
    \rowcolor{cat3}\multicolumn{15}{l}{Cross-modal ReID methods } \\
    \hline
     \rowcolor{cat3} & CM-NAS\cite{CM-NAS} & CVPR 2021 & 30.7 & 46.0 & 54.6 & 57.4 & 8.2 & 1.5 & 10.8 & 21.5 & 7.6 & 4.5 & 11.9 & 19.4 \\
     \rowcolor{cat3} & LbA \cite{lba} & CVPR 2021 & 33.0 & 48.3 & 59.7 & 62.5 & 11.9 & 4.6 & 23.1 & 41.5 & 8.5 & 6.0 & 14.9 & 22.4\\
     \rowcolor{cat3} & DEEN \cite{DEEN} & CVPR 2023 & 43.8 & 58.5 & 64.2 & 66.5 & 31.3 & 21.5 & 44.6 & 60.0 & 27.4 & 22.4 & 40.3 & 53.7\\
     \rowcolor{cat3} & MCJA \cite{MCJA} & TCSVT 2024 & 47.1 & 59.1 & 67.9 & 73.0 & 18.6 & 10.8 & 27.7 & 38.5 & 19.7 & 14.9 & 28.3 & 43.3\\
     \rowcolor{cat3} & VersReID \cite{versReID} & TPAMI 2024 & 49.3 & 59.7 & 70.5 & 78.4 & 25.7 & 13.8 & 40.0 & 61.5 & 27.7 & 17.9 &44.8 & 61.2\\
     \rowcolor{cat3} & AMML \cite{amml} & IJCV 2025 & 31.2 & 43.8 & 52.8 & 56.8 & 9.1 & 4.6 & 10.8 & 21.5 & 9.2 & 4.5 & 13.4 & 20.9\\
     \rowcolor{cat3} \multirow{-7}{*}{\begin{tabular}{@{}c@{}}Visible-Infrared \\ person ReID\end{tabular}} & HSFLNet \cite{hsflnet} & EAAI 2025 & 22.8 & 29.0 & 44.9 & 52.3 & 19.0 & 13.9 & 24.6 & 32.3 & 19.2 & 19.4 & 26.9 & 46.3\\
     \hline
     \rowcolor{cat3} & TransOSS \cite{hoss} & ICCV 2025 & \underline{57.4} & \underline{65.9} & \underline{79.5} & \underline{85.8} & \underline{48.9} & \underline{33.8} & \underline{67.7} & \textbf{80.0} & \underline{38.7} & \underline{29.9} & \underline{59.7} & \underline{71.6}\\
     \rowcolor{cat3} \multirow{-2}{*}{\begin{tabular}{@{}c@{}}Optical-SAR \\ ship ReID\end{tabular}} & MOS(ours) & - &  \textbf{60.4} & \textbf{68.8}  & \textbf{83.5} & \textbf{88.7} & \textbf{51.4} & \textbf{40.0} & \textbf{70.8} & \underline{75.4} & \textbf{48.7} & \textbf{46.3} & \textbf{68.7} & \textbf{79.1} \\
    
    \bottomrule
  \end{tabular}
\end{table*}

\subsection{Cross-modal data generation and feature fusion}

To further reduce the modality gap, we train a Brownian Bridge Diffusion Model (BBDM) \cite{li2023bbdm} to generate corresponding SAR images from the optical samples, and then fuse the features of both modalities for enhanced representation.

Following the BBDM formulation, we denote $x_0 \sim p_\text{SAR}(x)$ as a latent feature of a SAR sample, and $y \sim p_\text{opt}(y)$ as its corresponding optical feature. The forward process gradually transforms $x_0$ into $y$ through a Brownian bridge:
\begin{equation}
\begin{split}
q(x_t \mid x_0,  y)& = \mathcal{N}\Big(x_t;(1-m_t)x_0 + m_ty, \delta_t I \Big), \\
& m_t = \frac{t}{T}, \delta_t = 2(m_t - m_t^2)
\end{split}
\end{equation}
where $x_t$ denotes the noisy intermediate state at time $t$, conditioned on the endpoint $y$.

The reverse process predicts $x_{t-1}$ from $x_t$ and $y$ to gradually recover $x_0$:
\begin{equation}
p_\theta(x_{t-1} \mid x_t, y) = \mathcal{N}\Big(x_{t-1}; \mu_\theta(x_t, t, y), \hat{\delta}_tI \Big),
\end{equation}
where $\mu_\theta(x_t, t, y)$ is the predicted mean value of the noise, and $\hat{\delta}_t$ is the variance on noise at each step, and $\theta$ represents the model parameters.


The model is trained to predict the injected noise $\epsilon$ in the forward process. Specifically, given $x_t = (1 - m_t) x_0 + m_t y + \sqrt{\delta_t} \epsilon$ with $\epsilon \sim \mathcal{N}(0, I)$, we minimize the denoising objective:
\begin{equation}
\label{eq:loss_diff}
\mathcal{L}_{\text{diff}} = \mathbb{E}_{x_0, y, t, \epsilon} \left[ \big\| \epsilon - \epsilon_\theta(x_t, t, y) \big\|^2 \right],
\end{equation}
where $\epsilon_\theta(\cdot)$ is a neural network that estimates the noise given the noisy input $x_t$, time step $t$, and target optical feature $y$.


At inference, for a given optical sample $(I_i, y_i, \text{opt})$, we generate $K$ diverse pseudo-SAR samples $\{(\hat{I}_i^k, y_i, \text{sar})\}_{k=1}^K$. This enables the model to capture multiple plausible SAR representations corresponding to the same optical input, accounting for modality ambiguity and scene variability. Each generated pseudo-SAR image $\hat{I}_i^k$ and the original optical image $I_i$ is passed through the shared backbone network in \cref{sec3.2}, yielding $d$-dimensional features: $f_{\text{opt}}^i$ and $f_{\text{pseudo-sar}}^{i,k}, k = 1, \dots, K$.

The final fused representation is obtained by aggregating the optical feature with the ensemble of generated SAR features through a weighted combination:


\begin{equation}
f_{\text{fused}}^i = \frac{
    (1 - \tau) \, f_{\text{opt}}^i + \tau \left( \frac{1}{K} \sum_{k=1}^K f_{\text{pseudo-sar}}^{i,k} \right)
}{
    \left\| 
        (1 - \tau) \, f_{\text{opt}}^i + \tau \left( \frac{1}{K} \sum_{k=1}^K f_{\text{pseudo-sar}}^{i,k} \right)
    \right\|_2
},
\end{equation}

where $\tau \in [0, 1]$ controls the relative importance of the generated SAR ensemble. 
This weighting scheme allows the model to adaptively balance between the reliable but modality-specific optical feature and the diverse but potentially noisy synthetic SAR features. 
In our experiments, we set $K=5$ and treat $\tau$ as a hyperparameter, selecting its value on the validation set to optimize retrieval performance.

\section{Experiments}
\subsection{Dataset and metrics}

\begin{table}
  \caption{Detailed information about the test set HOSS ReID datasets}
  \label{table:dataset}
  \centering
  \setlength{\tabcolsep}{4.pt}
  \begin{tabular}{@{}c|ccc|ccc@{}}
    \toprule
    \multirow{2}{*}{Protocol} & \multicolumn{3}{c|}{Query images} & \multicolumn{3}{c}{Gallery images} \\
     & Optical & SAR & All & Optical & SAR & All \\
    \midrule
     \textit{All to ALL} & 88 & 88 & 176 & 403 & 190 & 593 \\
     \textit{Optical to SAR} & 65 & 0 & 65 & 0 & 190 & 190 \\
     \textit{SAR to Optical} &  0 & 67 & 67 & 403 & 0 & 403 \\

    \bottomrule
  \end{tabular}
\end{table}

To the best of our knowledge, there exists only a single publicly accessible dataset, namely HOSS ReID \cite{hoss}, for optical-SAR ship reID. Its training set contains 1,063 images, including 574 in optical modality and 489 in SAR modality. Detailed information test set about HOSS ReID is shown in \cref{table:dataset}. We evaluate all methods using standard retrieval metrics: mean Average Precision (mAP), and Rank-\{$1$, $5$, $10$\} accuracy (R1, R5, R10). 

\subsection{Implementation details}
\label{sec4.2}

We adopt TransOSS~\cite{hoss} as our baseline and follow its default configuration, using a learning rate of $5 \times 10^{-4}$ and batch size of 32, and setting both $\lambda_{\mathrm{id}}$ and $\lambda_{\mathrm{tri}}$ to 1.0. 
For MCRL, we train the model for 100 epochs, setting $\alpha = 5.0$ to strengthen the SAR image denoising process and scaling $\lambda_{\mathrm{cmal}}$ to 2.0 to balance the optimization across loss terms.

For CDGF, we pre-train the BBDM generator~\cite{li2023bbdm} for 100 epochs on the QXS-SAROPT dataset~\cite{huang2021qxs}, then fine-tune it for an additional 250 epochs on the HOSS ReID training set~\cite{hoss}. Given the limited size of the HOSS ReID training set, we apply data augmentation by horizontally and vertically flipping a subset of optical and SAR samples to increase diversity and improve model generalization. During inference, the fusion weight $\tau$ is set to 0.2 to prioritize the real optical features while incorporating complementary structural information from the generated pseudo-SAR samples. 
All experiments are conducted on a single NVIDIA A100-SXM4-80GB GPU.
\begin{table}
  \caption{Ablation study on the HOSS ReID dataset. Bold values indicate the best performance. "\textit{O to S}" means "\textit{Optical to SAR}" and "\textit{S to O}" means "\textit{SAR to Optical}".}
  \label{ablation}
  \centering
  \setlength{\tabcolsep}{4pt}
  \begin{tabular}{@{}cc|cc|cc|cc@{}}
    \toprule
    \multirow{2}{*}{MCRL} & \multirow{2}{*}{CDGF} & \multicolumn{2}{|c|}{\textit{ALL to ALL}} & \multicolumn{2}{c|}{\textit{O to S}} & \multicolumn{2}{c}{\textit{S to O}} \\
     & & mAP & R1 & mAP & R1 & mAP & R1 \\ 
     \midrule
     \ding{55} & \ding{55} & 57.4 & 65.9 & 48.9 & 33.8 & 38.7 & 29.9 \\
     \checkmark & \ding{55} & 59.3 & 68.2 & 50.3 & 38.5 & 45.9 & 40.3\\
     \ding{55} & \checkmark & 58.1 & 66.5 & 49.7 & 36.9 & 38.6 & 32.8 \\
     \checkmark & \checkmark & \textbf{60.4} & \textbf{68.8} & \textbf{51.4} & \textbf{40.0} & \textbf{48.7} & \textbf{46.3} \\
    \bottomrule
  \end{tabular}
\end{table}
\begin{figure}[t]
  \centering
   \includegraphics[width=1.0\linewidth]{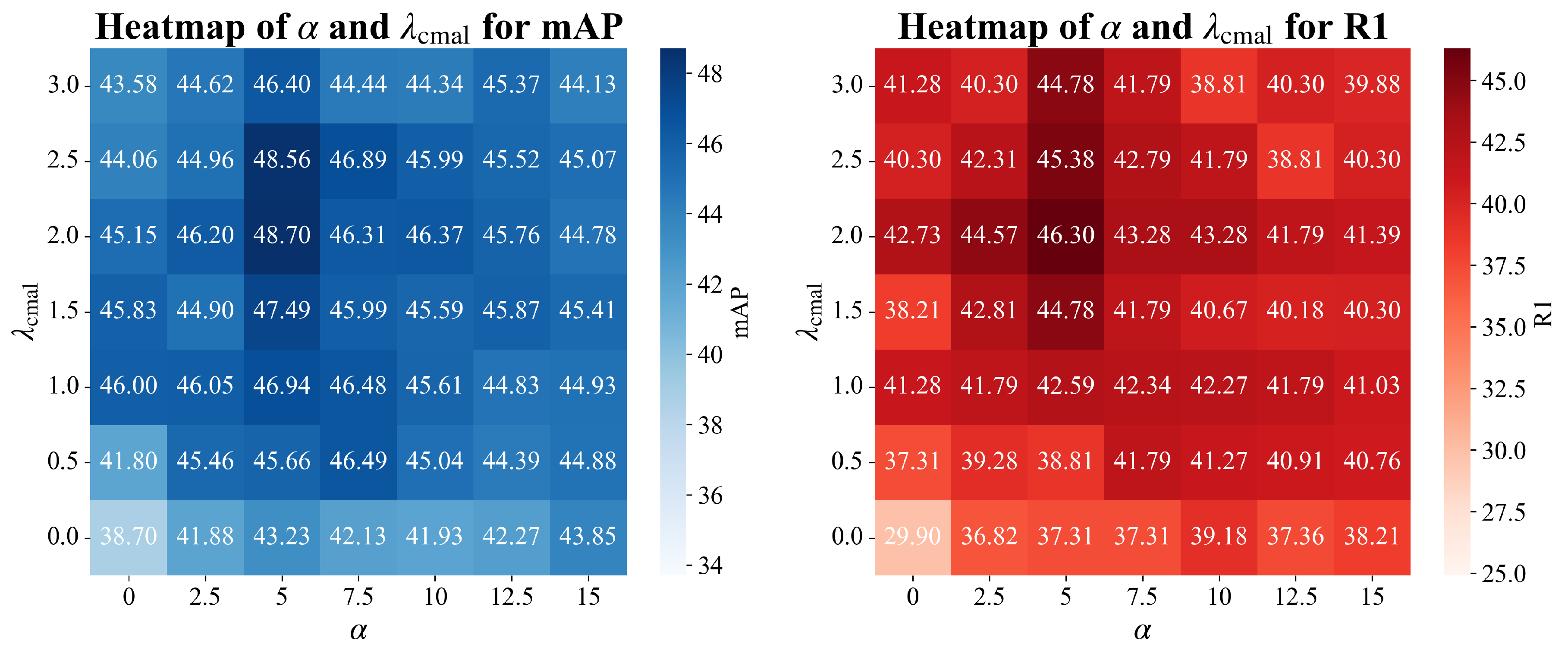}
   \caption{R1 heatmap based on $\alpha$ and $\lambda_\text{cmal}$ in \textit{SAR to Optical} protocol.}
   \label{fig:heatmap}
\end{figure}
\subsection{Comparison with state-of-the-art methods}
To the best of our knowledge, optical–SAR cross-modal ship ReID remains an emerging and underexplored task, with few open-source methods \cite{hoss} tailored for this domain. Therefore, we evaluate several state-of-the-art person ReID approaches on the HOSS ReID dataset. The results are summarized in \cref{results}, where “\textit{ALL to ALL}” denotes mixed-modality retrieval, “\textit{Optical to SAR}” represents optical-to-SAR retrieval, and “\textit{SAR to Optical}” indicates the reverse.

We divide the compared methods into three categories: general vision methods, single-modality ReID methods, and cross-modal ReID methods. Experimental results show that visible–infrared ReID methods \cite{CM-NAS, lba, DEEN, MCJA, versReID, amml, hsflnet} perform the worst across all protocols, mainly due to the inherent discrepancy between infrared and SAR imaging mechanisms. Their architectures, originally optimized for infrared feature extraction, fail to capture discriminative representations from SAR data. In contrast, ship ReID methods \cite{D2InterNet, hoss} outperform person ReID methods \cite{SOLIDER, transreid}, as they better model the structural and appearance characteristics of ships. Excluding our proposed MOS, TransOSS \cite{hoss} attains the highest performance among existing methods, benefiting from its task-specific design tailored for HOSS ReID. More importantly, MOS consistently outperforms TransOSS~\cite{hoss} across all protocols. 
In \textit{ALL to ALL} protocol, MOS achieves 60.4\% mAP (+3.0\%) and 68.8\% R1 (+2.9\%); in \textit{Optical to SAR} protocol, 51.4\% mAP (+2.5\%) and 40.0\% R1 (+6.2\%); and in \textit{SAR to Optical} protocol, 48.7\% mAP (+10.0\%) and 46.3\% R1 (+16.4\%), demonstrating superior generalization and modality bridging.
\begin{figure}[t]
  \centering
   \includegraphics[width=0.9\linewidth]{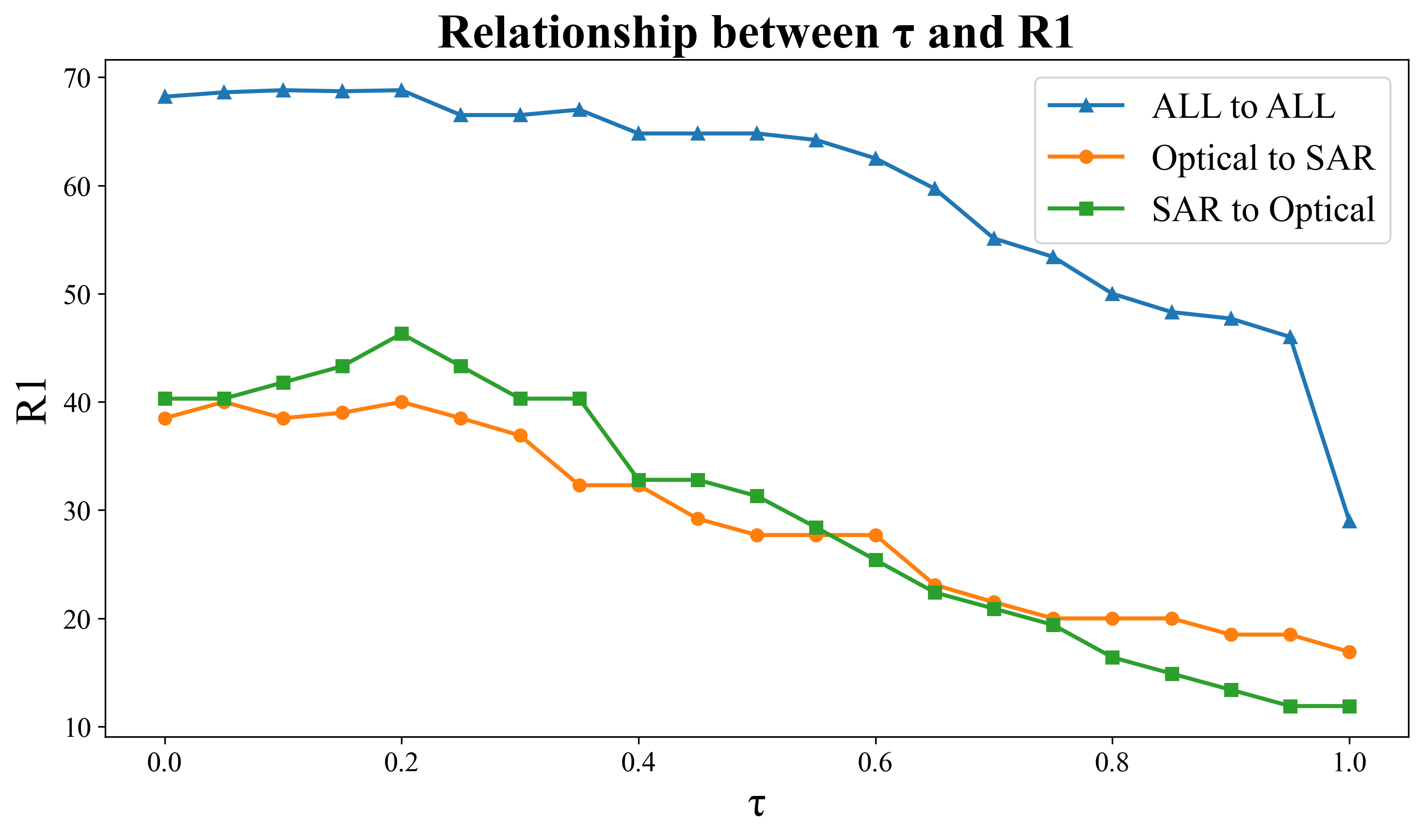}

   \caption{Visualization of relationship of $\tau$ and R1 in different evaluation protocols.}
   \label{fig:tau_vs_r1}
\end{figure}
\subsection{Ablation study}
We conduct an ablation study on the HOSS ReID dataset to evaluate the proposed MCRL and CDGF modules, with results summarized in \cref{ablation}.  
The addition of MCRL boosts performance by +1.9\% mAP and +2.3\% R1, yielding substantial improvements in the challenging \textit{SAR to Optical} protocol: +7.2\% mAP and +10.4\% R1. This confirms that MCRL effectively reduces the modality gap during training by enforcing modality-consistent feature learning. In contrast, CDGF alone yields only marginal improvements, as its inference-time feature fusion relies on a well-aligned embedding space. Without MCRL to regularize the backbone, the large cross-modal discrepancy limits its effectiveness.
\begin{figure*}[t]
  \centering
   \includegraphics[width=0.8\linewidth]{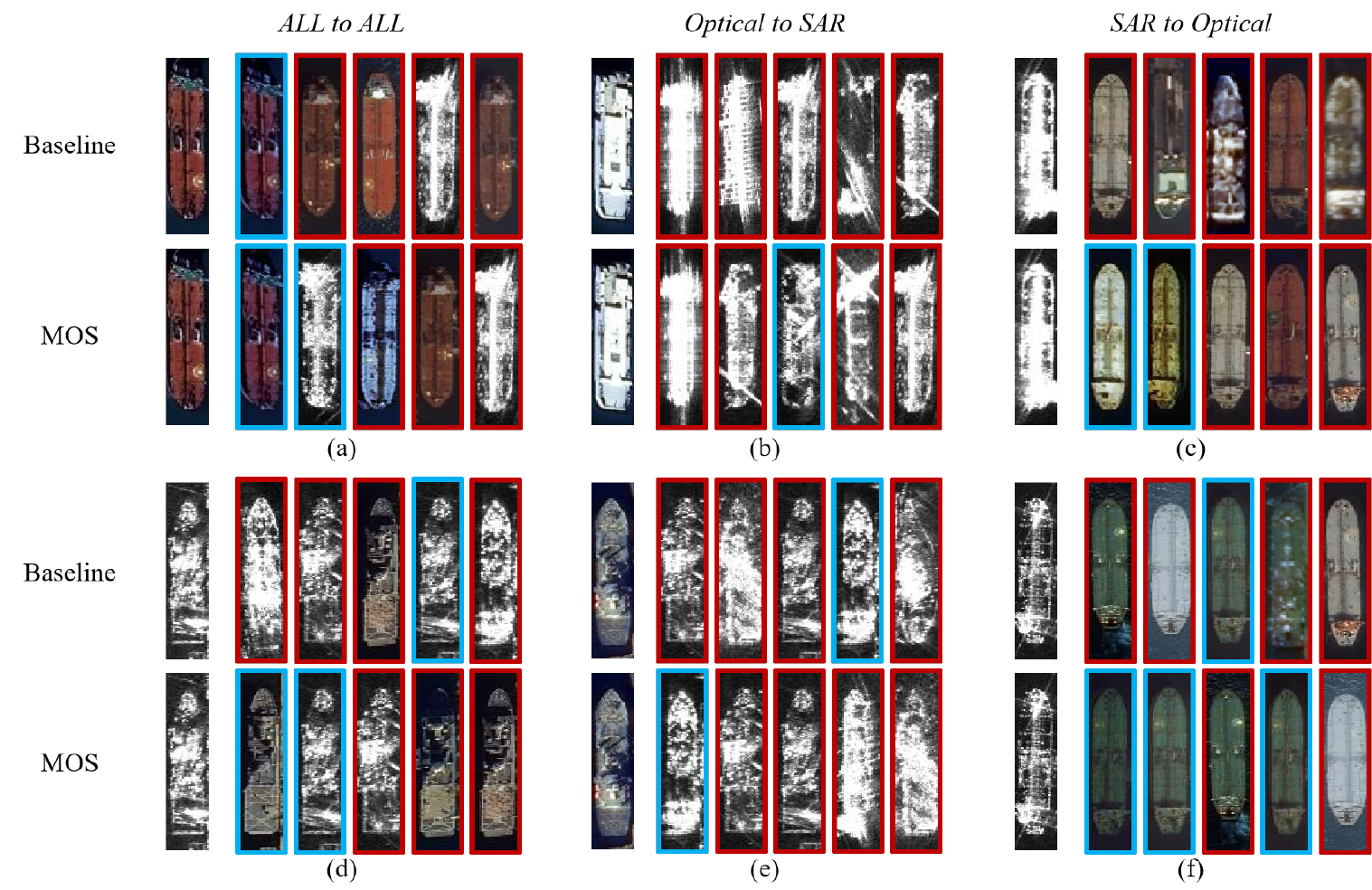}

   \caption{Visualization of top-5 retrieval results in \textit{ALL to ALL}, \textit{Optical to SAR} and \textit{SAR to Optical} protocols. Blue borders indicate correct search results, while red borders denote errors.}
   \label{fig:vis}
\end{figure*}
When both modules are combined, we achieve the best performance: +3.0\% and +2.9\% in \textit{ALL to ALL}, +2.5\% mAP and +6.2\% R1 in \textit{Optical to SAR}, and +10.0\% mAP and +16.4\% R1 in \textit{SAR to Optical} protocol. The results demonstrate that MCRL and CDGF are complementary: MCRL establishes a robust, aligned feature space during training, while CDGF enhances inference-time consistency through cross-modal feature fusion. This two-stage alignment, combining early regularization with late fusion, proves particularly effective for extreme modality shifts.

\subsection{Hyperparameter sensitivity analysis}
We fix $\tau = 0.2$ and analyze the effect of $\alpha$ and $\lambda_{\text{cmal}}$ in \textit{SAR to Optical} protocol (other protocols in supplementary material). 
As shown in \cref{fig:heatmap}, the optimal performance is achieved when $\alpha = 5.0$ and $\lambda_{\text{cmal}} = 2.0$ across all protocols. 
When $\alpha$ is too small, noise is insufficiently suppressed, leading to corrupted feature learning; when $\alpha$ is too large, valuable semantic pixels are over-removed, degrading the quality of clean pseudo-labels. 
Similarly, extreme values of $\lambda_{\text{cmal}}$ disrupt the balance between the classification and alignment losses, causing performance degradation. 

We fix all other hyperparameters to the values set in \cref{sec4.2}, while analyzing the effect of $\tau$ across different evaluation protocols. As shown in \cref{fig:tau_vs_r1}, all three protocols achieve near-optimal performance when $\tau = 0.2$, with R1 scores reaching or closely approaching their maximum values. 
This suggests that $\tau = 0.2$ provides an effective balance for feature fusion. Based on this observation, we adopt $\tau = 0.2$ as the default value in all experiments.
\subsection{Visualization}

We present qualitative top-5 retrieval results in \cref{fig:vis} to illustrate the effectiveness of our MOS framework. Each row displays the top-5 retrieved samples for a given query, with blue borders indicating correct matches and red borders denoting incorrect ones.

MOS demonstrates significantly improved retrieval performance compared to the baseline. For instance, as shown in (c), when the baseline fails to retrieve the correct match within the top-5 results, MOS successfully ranks the correct sample at the first position. In (d), MOS advances the ranking of correct matches compared to the baseline, leading to improved retrieval performance. In (f), while the baseline retrieves the correct match only at the third position, MOS successfully recovers all three correct samples from the gallery, ranked at positions one, two, and four (although these three sample are visually similar, they are distinct, \eg, slight differences in viewing angle or imaging conditions), demonstrating MOS's superior recall and robustness in challenging cross-modal matching.

Moreover, in the \textit{ALL to ALL} setting, MOS overcomes the baseline's tendency to retrieve samples from only a single modality due to large cross-modal discrepancies. Instead, MOS enables flexible and balanced retrieval across both two modalities. As demonstrated in (a) and (d), the retrieved results exhibit a well-mixed distribution of modalities, with the top-ranked matches (first and second positions) coming from different modalities but sharing the same identity, indicating that MOS achieves true cross-modal generalization without bias toward either modality.

\section{Conclusion}

In this paper we propose MOS, a novel framework for Mitigating Optical-SAR modality gap for cross-modal ship ReID. At the core of MOS are two complementary modules: Modality-Consistency Representation Learning (MCRL) and Cross-Modal Data Generation and Feature fusion (CDGF). 
MCRL reduces the modality gap during training by aligning the modality distribution, while simultaneously enhancing SAR feature quality through implicit denoising. 
CDGF further improves inference-time robustness by generating pseudo-SAR samples from optical inputs and fusing their features with the original representations, thereby promoting cross-modal consistency. 
Extensive experiments on the HOSS ReID dataset, the only publicly available benchmark for this task, demonstrate that MOS achieves state-of-the-art performance, outperforming existing methods across all evaluation protocols.
\clearpage
{
    \small
    \bibliographystyle{ieeenat_fullname}
    \bibliography{main}
}
\clearpage
\maketitlesupplementary

\section{Wasserstein-2 distance}

The Wasserstein-2 distance ($W_2$) is a fundamental metric from optimal transport that measures the minimal effort required to transform one probability distribution into another. For two distributions $p$ and $q$ defined on a metric space $\mathcal{X}$, the $W_2$ distance is given by:
\begin{equation}
W_2(p,q) = 
\left(
\inf_{\gamma \in \Gamma(p,q)}
\mathbb{E}_{(x,y)\sim\gamma}\left[\|x-y\|_2^2\right]
\right)^{1/2},
\end{equation}
where $\Gamma(p,q)$ denotes the set of couplings whose marginals are $p$ and $q$. The infimum corresponds to the optimal transport plan that minimizes the quadratic transportation cost.

A key advantage of $W_2$ is that it admits a closed-form expression when both distributions are Gaussian. Let
\[
p = \mathcal{N}(\mu_1,\Sigma_1), \qquad
q = \mathcal{N}(\mu_2,\Sigma_2),
\]
then the squared Wasserstein-2 distance is:
\begin{equation}
\begin{split}
W_2^2(p,q)
= {} & \|\mu_1 - \mu_2\|_2^2  \\
& + \mathrm{Tr}\!\Big(
\Sigma_1 + \Sigma_2 
- 2\,(\Sigma_2^{1/2}\Sigma_1 \Sigma_2^{1/2})^{1/2}
\Big).
\end{split}
\end{equation}

This formulation shows that $W_2$ jointly aligns the means and covariances of the two distributions, providing a geometry-aware and numerically stable measurement of distributional discrepancy. Compared with KL divergence or simple $\ell_2$ matching, $W_2$ remains finite for distributions with disjoint support and captures covariance differences naturally. These properties make it particularly effective for aligning Gaussian feature distributions in our method.

\begin{figure}[t]
  \centering
   \includegraphics[width=1.0\linewidth]{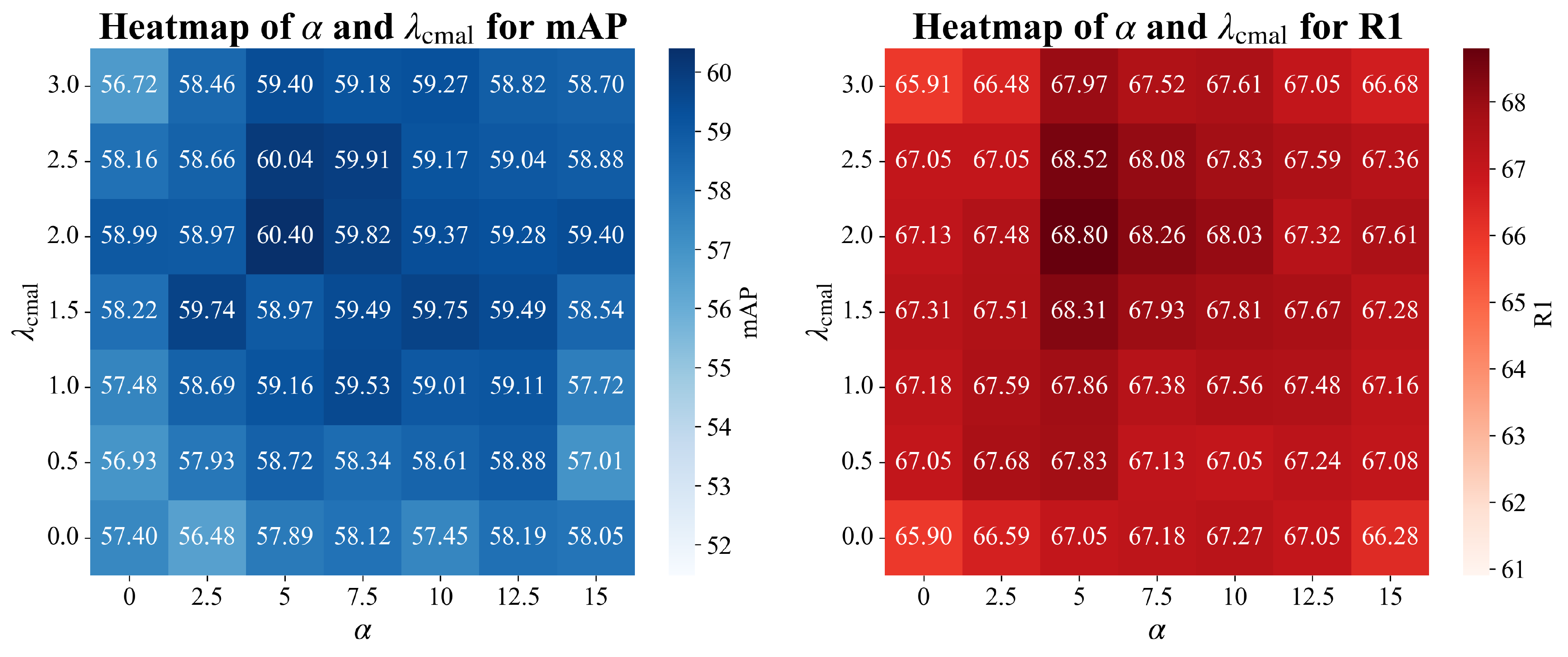}

   \caption{R1 heatmap based on $\alpha$ and $\lambda_\text{cmal}$ in \textit{ALL to ALL} protocol.}
   \label{fig:heatmap_supp_all}
\end{figure}

\begin{figure}[t]
  \centering
   \includegraphics[width=1.0\linewidth]{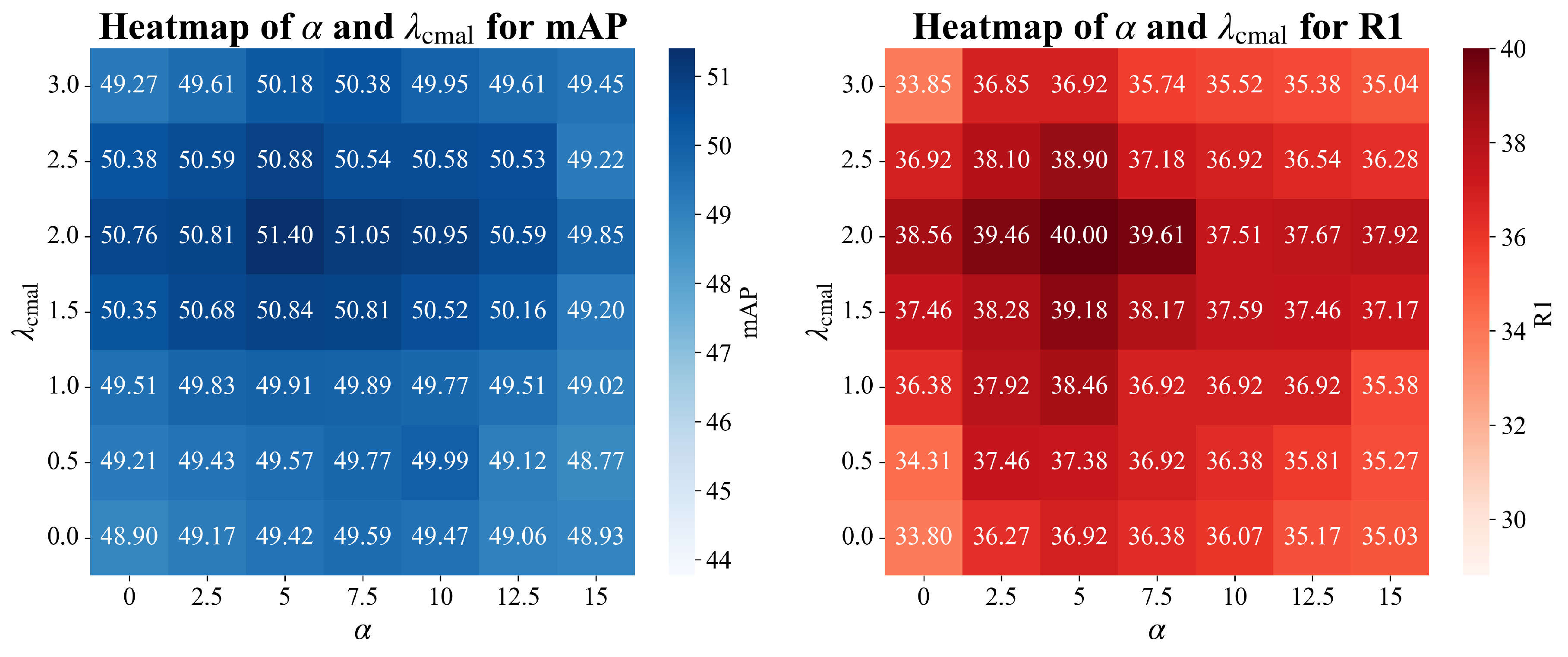}

   \caption{R1 heatmap based on $\alpha$ and $\lambda_\text{cmal}$ in \textit{Optical to SAR} protocol.}
   \label{fig:heatmap_supp_o2s}
\end{figure}



\section{Heatmap in other evaluation protocols}

Both heatmaps in Fig.~\ref{fig:heatmap_supp_all} and Fig.~\ref{fig:heatmap_supp_o2s} illustrate the sensitivity of our method with respect to the hyper-parameters $\alpha$ and $\lambda_\text{cmal}$ under the \textit{ALL to ALL} and \textit{Optical to SAR} protocols. Despite the differences in evaluation settings, the two heatmaps exhibit highly consistent trends: the optimal performance is achieved when $\alpha = 5.0$ and $\lambda_{\text{cmal}} = 2.0$ across all protocols. This consistency indicates that our cross-modal alignment strategy is robust and not overly sensitive to the choice of $\alpha$ and $\lambda_\text{cmal}$ across different protocols. 



\end{document}